\documentclass[10pt]{article} 
\usepackage{times}

\usepackage[numbers]{natbib}

\usepackage{graphicx} 
\usepackage{subfigure}

\usepackage{synttree}

\usepackage{algorithm}
\usepackage{algorithmic}

\usepackage{comment}
\usepackage{amsmath,amsthm,amsfonts}
\usepackage{multirow}
\usepackage{url}

\graphicspath{{figs/}}

\newcommand{\xM}{\bar{x}}
\newcommand{\yM}{\bar{y}}

\title{Textual Features for Programming by Example}

\author{
{Aditya Krishna Menon} \\
{University of California, San Diego,}\\
{9500 Gilman Drive, La Jolla CA 92093}\\
\url{akmenon@ucsd.edu}
\and
{Omer Tamuz}\\
{Faculty of Mathematics and Computer Science,}\\
{The Weizmann Institute of Science, Rehovot Israel}\\
\url{omer.tamuz@weizmann.ac.il}
\and
{Sumit Gulwani}\\
{Microsoft Research,}\\
{One Microsoft Way, Redmond, WA 98052}\\
\url{sumitg@microsoft.com}
\and
{Butler Lampson}\\
{Microsoft Research,}\\
{One Memorial Drive, Cambridge MA 02142} \\
\url{butler.lampson@microsoft.com}
\and
{Adam Tauman Kalai}\\
{Microsoft Research,}\\
{One Memorial Drive, Cambridge MA 02142}\\
\url{adum@microsoft.com}
}

\begin{document}

\date{\today}
\maketitle

\begin{abstract}
In Programming by Example, a system attempts to infer a program from input and output examples, generally by searching for a composition of certain base functions. Performing a na\"{i}ve brute force search is infeasible for even mildly involved tasks. We note that the examples themselves often present {\em clues} as to which functions to compose, and how to rank the resulting programs. In text processing, which is our domain of interest, clues arise from simple \emph{textual features}: for example, if parts of the input and output strings are permutations of one another, this suggests that sorting may be useful. We describe a system that \emph{learns} the reliability of such clues, allowing for faster search and a principled ranking over programs. Experiments on a prototype of this system show that this learning scheme facilitates efficient inference on a range of text processing tasks.
\end{abstract}

\section{Introduction}

Programming by Example (PBE)~\citep{lieberman, Cypher} is an attractive means for end-user programming tasks, wherein the user provides the machine \emph{examples} of a task she wishes to perform, and the machine infers a program to accomplish this. This paradigm has been used in a wide variety of domains; \citep{wambse12} gives a recent overview. We focus on \emph{text processing}, a problem most computer users face (be it reformatting the contents of an email or extracting data from a log file), and for which several complete PBE systems have been designed, including LAPIS \citep{lapis}, SMARTedit \citep{lau-icml00}, QuickCode \citep{Gulwani,vldb12}, and others~\citep{Nix85,TELS}. Such systems aim to provide a simpler alternative to the traditional solutions to the problem, which involve either tedious manual editing, or esoteric computing skills such as knowledge of {\tt awk} or {\tt emacs}.

A fundamental challenge in PBE is the following \emph{inference} problem: given a set of base functions, how does one quickly search for programs composed of these functions that are consistent with the user-provided examples? One way is to make specific assumptions about the nature of the base functions, as is done by many existing PBE systems \cite{lapis,lau-icml00,Gulwani}, but this is unsatisfying because it restricts the range of tasks a user can perform. The natural alternative, brute force search, is infeasible for even mildly involved programs \cite{dimensions}. Thus, a basic question is whether there is a solution possessing both generality and efficiency.

This paper aims to take a step towards an affirmative answer to this question. We observe that there are often telling {\em features} in the user examples suggesting which functions are likely. For example, suppose that a user demonstrates their intended task through one or more input-output pairs of strings $\{(x_i,y_i)\}$, where each $y_i$ is a permutation of $x_i$. This feature provides a {\em clue} that when the system is searching for the $f(\cdot)$ such that $f(x_i)=y_i$, sorting functions may be useful. Our strategy is to incorporate a library of such {clues}, each suggesting relevant functions based on textual features of the input-output pairs. We \emph{learn} weights telling us the reliability of each clue, and use this to bias program search. This bias allows for significantly faster inference compared to brute force search. Experiments on a prototype system demonstrate the effectiveness of feature-based learning.

To clarify matters, we step through a concrete example of our system's operation.

\subsection{Example of our system's operation}
\label{sec:example-operation}

\begin{figure}[!htb]
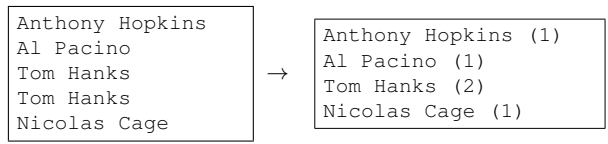

	\centering
	\footnotesize
	{\ttfamily
		\fbox{\parbox{0.25\linewidth}{
		Anthony Hopkins\\
		Al Pacino\\
		Tom Hanks\\
		Tom Hanks\\
		Nicolas Cage}}
		$\rightarrow$
		\setlength\fboxrule{0.3pt}
		\fbox{\parbox{0.3\linewidth}{
		Anthony Hopkins (1)\\
		Al Pacino (1) \\
		Tom Hanks (2)\\
		Nicolas Cage (1)}}
	}	
	\caption{ Input-output example for the desired task.
	\label{fig:conf}
	}
\end{figure}

Imagine a user has a long list of names with some repeated entries (say, the Oscar winners for Best Actor), and would like to create a list of the unique names, each annotated with their number of occurrences. Following the PBE paradigm
, in our system, the user illustrates the operation by providing an example, which is an input-output pair of strings. Figure \ref{fig:conf} shows one possible such pair, which uses a subset of the full list (in particular, the winners from '91--'95) the user possesses.

One way to perform the above transformation is to first generate an intermediate list where each element of the input list is appended with its occurrence count -- which would look like {\footnotesize \tt[ "Anthony Hopkins (1)", "Al Pacino (1)", "Tom Hanks (2)", "Tom Hanks (2)", "Nicolas Cage (1)"]} -- and then remove duplicates. The corresponding program $f(\cdot)$ may be expressed as the composition
$$f(x) = {\tt dedup}({\tt concatLists}(x, ``\ ", {\tt concatLists}(``(", {\tt count}(x, x), ``)"))).$$
The argument $x$ here represents the list of input lines that the user wishes to process, which may be much larger than the input provided in the example. We assume here a base language comprising (among others) a function {\tt\small dedup} that removes duplicates from a list, {\tt\small concatLists} that concatenates lists of strings elementwise, implicitly expanding singleton arguments, and {\tt\small count} that finds the number of occurrences of the elements of one list in another. 

While simple, this example is out of scope for existing text processing PBE systems. Most systems support a restricted, pre-defined set of functions that do not include natural tasks like removing duplicates; for example \cite{Gulwani} only supports functions that operate on a line-by-line basis. These systems perform inference with search routines that are hand-coded for their supported functionality, and are thus not easily extensible. (Even if an exception could be made for specific examples like the one above, there are countless other text processing applications we would like to solve.) Systems with richer functionality are inapplicable because they perform inference with brute force search (or a similarly intractable operation \cite{icse10}) over all possible function compositions. Such a na\"{i}ve search over even a moderate sized library of base functions is unlikely to find the complex composition of our example. Therefore, a more generic framework is needed.

Our basic observation is that certain textual features can help bias our search by providing clues about which functions may be relevant: in particular, (a) there are duplicate lines in the input but not output, suggesting that {\small\tt dedup} may be useful, (b) there are parentheses in the output but not input, suggesting the function {\small\tt concatLists("(",}$L${\tt ,")")} for some list $L$, (c) there are numbers on each line of the output but none in the input, suggesting that {\small\tt count} may be useful, and (d) there are many more spaces in the output than the input, suggesting that {\tt " "} may be useful. Our claim is that by \emph{learning} weights that tell us the reliability of these clues -- for example, how confident can we be that duplicates in the input but not the output suggests {\small\tt dedup} -- we can significantly speed up the inference process over brute force search. 

In more detail, a clue is a function that generates rules in a probabilistic context free grammar based on features of the provided example. Each rule corresponds to a function\footnote{When we describe clues as suggesting functions, we implicitly mean the corresponding grammar rule.} (possibly with bound arguments) or constant in the underlying programming language. The rule probabilities are computed from weights on the clues that generate them, which in turn are learned from a training corpus of input-output examples. To learn $f(\cdot)$, we now search through derivations of this grammar in order of decreasing probability. Table \ref{tbl:grammar} illustrates what the grammar may look like for the above example. Note that the grammar rules and probabilities are \emph{example specific}; we do not include a rule such as {\small\tt DELIM$\rightarrow$ "\$"}, say, because there is no instance of {\small\tt "\$"} in the input or output. Further, compositions of rules may also be generated, such as {\small\tt concatList("(",LIST,")")}.
\begin{table}[htb]
	\centering
	{\small
	\begin{tabular}{|ll|ll|}
	\hline
	\textbf{Production} & \textbf{Probability} & \textbf{Production} & \textbf{Probability} \\
	\hline
	{\tt P$\rightarrow$join(LIST,DELIM)} & 1 & {\tt CAT$\rightarrow$LIST}&0.7	 \\
	{\tt LIST$\rightarrow$split(x,DELIM)}&0.3&{\tt CAT$\rightarrow$DELIM}&0.3	\\
	{\tt LIST$\rightarrow$concatList(CAT,CAT,CAT)}&0.1& {\tt DELIM$\rightarrow$"$\backslash$n" }&0.5	\\
	{\tt LIST$\rightarrow$concatList("(",CAT,")")}&0.2&{\tt DELIM$\rightarrow$" "}&0.3\\
	{\tt LIST$\rightarrow$dedup(LIST)}&0.2&	{\tt DELIM$\rightarrow$"("}&0.1  \\
	{\tt LIST$\rightarrow$count(LIST,LIST)}& 0.2 & {\tt DELIM$\rightarrow$")"} & 0.1 \\
	\hline
	\end{tabular}
	}
	\caption{Example of grammar rules generated for task in Figure \ref{fig:conf}.}
	\label{tbl:grammar}
	\vspace{-2mm}
\end{table}

Table \ref{tbl:grammar} is of course a condensed view of the actual grammar our prototype system generates, which is based on a large library of about $100$ features and clues. With the full grammar, a na\"{i}ve brute force search over compositions takes $30$ seconds to find the right solution to the example of Figure \ref{fig:conf}, whereas with learning the search terminates in just $0.5$ seconds.

\subsection{Contributions}

To the best of our knowledge, ours is the first PBE system to exploit textual features for inference, which we believe is a step towards achieving the desiderata of efficiency and generality. The former will be demonstrated in an empirical evaluation of our learning scheme. For the latter, 
while the learning component is discussed in the context of text processing, the approach could possibly be adapted for different domains. Further, the resulting system is highly extensible. Through the use of clues, one only considers broadly relevant functions during the search for a suitable program: one is free to add functionality to process addresses, e.g.\,, without fear of it adversely affecting the performance of processing dates. Through the use of learning, we further sift amongst these broadly relevant functions, and determine which of them is likely to be useful in explaining the given data. A system designer need only write clues for any new functionality, and add relevant examples to the training corpus. Our system then automatically learns weights associated with these clues.

\subsection{Comparison to previous learning systems}

Most previous PBE systems for text processing handle a relatively small subset of natural text processing tasks. This is in order to admit efficient representation and  search over consistent programs, e.g.\ using a version space \citep{LauVersionSpace}, thus sidestepping the issue of searching for programs using general classes of functions. To our knowledge, every system designed for a library of arbitrary functions searches for appropriate compositions of functions either by brute force search, or a similarly intractable operation such as invoking a {\small\sf SAT} solver \cite{icse10}.\footnote{One could consider employing heuristic search techniques such as Genetic Programming. However, this requires picking a metric that captures meaningful search progress. This is difficult, since functions like sorting cause drastic changes on an input. Thus, standard metrics like edit distance may not be appropriate.} Our learning approach based on textual features is thus more general and flexible than previous approaches.

Having said this, our goal in this paper is \emph{not} to compete with existing PBE systems in terms of functionality. Instead, we wish to show that the fundamental PBE inference problem may be attacked by learning with textual features. This idea could in fact be applied in \emph{conjunction} with prior systems. A specific feature of the data, such as the input and output having the same number of lines, may be a clue that a function corresponding to a system like QuickCode \citep{Gulwani} will be useful.

\section{Formalism of our approach}
\label{sec:our-approach}

We begin a formal discussion of our approach by defining the learning problem in PBE.

\subsection{Programming by example (PBE)}

Let $\mathcal{S}$ denote the set of strings. At \emph{inference time}, the user provides a \emph{system input} $z := (x, \xM, \yM) \in \mathcal{S}^3$, where $x$ represents the data to be processed, and $(\xM, \yM)$ is the example input-output pair that represents the string transformation the user wishes to perform. In the example of the previous section, $(\xM, \yM)$ is the pair of strings represented in Figure \ref{fig:conf}, and $x$ is the list of all Oscar winners. While a typical choice for $\xM$ is some prefix of $x$, this is not required in general\footnote{This is more general than the setup of e.g.\ \citep{Gulwani}, which assumes $\xM$ and $\yM$ have the same number of lines, each of which is treated as a separate example.}. We assume that $\yM=f(\xM)$,
for some unknown {\em target function} or \emph{program} $f\in \mathcal{S}^\mathcal{S}$, from the set of functions that map strings to strings. Our goal is to recover $f(\cdot)$.

We do so by defining a probability model $\Pr[f | z; \theta]$ over programs, parameterized by some $\theta$. Given some $\theta$, at inference time on input $z$, we pick the most likely program under $\Pr[f | z; \theta]$ which is also consistent with $z$. We do so by invoking a {\em search function} $\sigma_{\theta, \tau}:\mathcal{S}^3\rightarrow \mathcal{S}^\mathcal{S}$ that depends on $\theta$ and an upper bound $\tau$ on search time. This produces our conjectured program $\hat{f}=\sigma_{\theta,\tau}(z)$ computing a string-to-string transformation, or a trivial failure function $\perp$ if the search fails in the allotted time.

The $\theta$ parameters are \emph{learned} at {training time}, where the system is given a corpus of $T$ training quadruples, $\{ (z^{(t)}, y^{(t)}) \}_{t=1}^T$, with $z^{(t)} = (x^{(t)}, \xM^{(t)}, \yM^{(t)}) \in \mathcal{S}^3$ representing the actual data and the example input-output pair, and $y^{(t)} \in \mathcal{S}$ the correct output on $x^{(t)}$. Note that each quadruple here represents a different \emph{task}; for example, one may represent the Oscar winners example of the previous section, another a generic email processing task, and so on. From these examples, the system chooses the parameters $\theta$ that maximize the likelihood $\Pr[f | z; \theta]$. We now describe how we model the conditional distribution $\Pr[f | z; \theta]$ using a probabilistic context-free grammar.

\subsection{PCFGs for programs}
\label{sec:pcfg}

We maintain a probability distribution over programs with a {\em Probabilistic Context-Free Grammar} (PCFG) $\mathcal{G}$, as discussed in \citep{Liang}. The grammar is defined by a set of non-terminal symbols $\mathcal{V}$, terminal symbols $\Sigma$ (which may include strings $s \in \mathcal{S}$ and also other program-specific objects such as lists or functions), and rules $\mathcal{R}$. Each rule $r \in \mathcal{R}$ has an associated probability $\Pr[r | z; \theta]$ of being generated given the system input $z$, where $\theta$ represents the unobserved parameters of the grammar.  WLOG, each rule $r$ is also associated with a function $f_r : {\Sigma}^{\text{NArgs}(r)} \to {\Sigma}$, where NArgs$(r)$ denotes the number of arguments in the RHS of rule $r$. 
A program\footnote{Two \emph{programs} from different derivations may compute exactly the same \emph{function} $f:\mathcal{S} \rightarrow\mathcal{S}$.  However, determining whether two programs compute the same function is undecidable in general.  Hence, we abuse notation and consider these to be different functions.} is a derivation of the start symbol $\mathcal{V}_\text{start}$. The probability of any program $f(\cdot)$ is the probability of its constituent rules $\mathcal{R}_f$ (counting repetitions):
\begin{align}
\label{eqn:PCFG}
\Pr[f | z; \theta] = \Pr[\mathcal{R}_f | z; \theta] = \prod_{r \in \mathcal{R}_f} \Pr[r | z; \theta].
\end{align}
We now describe how the distribution $\Pr[r | z; \theta]$ is parameterized using clues.

\subsection{Features and clues for learning}

The learning process exploits the following simple fact: the chance of a rule being part of an explanation for a string pair $(\xM, \yM)$ depends greatly on certain characteristics in the structure of $\xM$ and $\yM$. For example, one interesting binary {\em feature} is whether or not every line of $\yM$ is a substring of $\xM$.
If true, it may suggest that the {\tt select\_field} rule should receive higher probability in the PCFG, and hence will be combined with other rules more often in the search.  Another binary feature indicates whether or not ``Massachusetts'' occurs repeatedly as a substring in $\yM$ but not in $\xM$.  This suggests that a rule generating the string ``Massachusetts'' may be useful. Conceptually, given a training corpus, we would like to learn the relationship between such features and the successful rules. However, there are an infinitude of such binary features as well as rules (e.g.\ a feature and rule corresponding to every possible constant string), but of course limited data and computational resources. So, we need a mechanism to estimate the relationship between the two entities.

We connect features with rules via {\em clues}. A clue is a function $c:\mathcal{S}^3\rightarrow 2^\mathcal{R}$ that states, for each system input $z$, which subset of rules in $\mathcal{R}$ (the infinite set of grammar rules), may be relevant. This set of rules will be based on certain features of $z$, meaning that we search over compositions of instance-specific rules\footnote{As long as the functions generated by our clues library include a Turing-complete subset, the class of functions being searched amongst is always the Turing-computable functions, though having a good bias is probably more useful than being Turing complete.}. For example, one clue might return {\small$\{ \mathtt{E} \to \mathtt{select\_field(E,Delim,Int)}\}$} if each line of $\yM$ is a substring of $\xM$, and $\emptyset$ otherwise.  Another clue might recognize the input string is a permutation of the output string, and generate rules {\small$\{\mathtt{E} \to \mathtt{sort(E,COMP)},\mathtt{E} \to \mathtt{reverseSort(E,COMP)},\mathtt{COMP} \rightarrow \mathtt{alphaComp},\ldots\}$}, i.e., rules for sorting as well as introducing a nonterminal along with corresponding rules for various comparison functions. Note that a single clue can suggest a multitude of rules for different $z$'s (e.g.\ $\mathtt{E} \to s$ for every substring $s$ in the input), and ``common'' functions (e.g.\ concatenation of strings) may be suggested by multiple clues. 

We now describe our probability model that is based on the clues formalism.

\subsection{Probability model}

Suppose the system has $n$ clues $c_1,c_2,\ldots,c_n$. For each clue $c_i$, we keep an associated parameter $\theta_i \in \mathbb{R}$. Let $\mathcal{R}_z = \cup_{i=1}^n c_i(z)$ be the set of \emph{instance-specific} rules (wrt $z$) in the grammar. While the set of all rules $\mathcal{R}$ will be infinite in general, we assume there are a finite number of clues suggesting a finite number of rules, so that $\mathcal{R}_z$ is finite. For each rule $r \notin \mathcal{R}_z$, we take $\Pr[r | z] = 0$, i.e.\ a rule that is not suggested by any clue is disregarded. For each rule $r \in \mathcal{R}_z$, we use the probability model
\begin{equation}\label{eqn:product}
\Pr[r~|~z;\theta] = \frac{1}{Z_{\mathrm{LHS}(r)}}  \exp\left(\sum_{i: r \in c_i(z)} \theta_i \right).
\end{equation}
where for each nonterminal $V$, the normalizer $Z_V$ ensures we get a valid probability distribution:
$$Z_V = \sum_{r \in \mathcal{R}_z: \mathrm{LHS}(r)=V} \exp\left(\sum_{i: r \in c_i(z)} \theta_i \right).$$
This is a log-linear model for the probabilities, where each clue has a {\em weight} $e^{\theta_i}$, which is intuitively its \emph{reliability}, and the probability of each rule is proportional to the {\em product} of the weights generating that rule. An alternative would be to make the probabilities be the (normalized) {\em sums} of corresponding weights, but we favor products for two reasons.  First, as described shortly, maximizing the log-likelihood is a convex optimization problem in $\theta$ for products, but not for sums.  Second, this formalism allows clues to have positive, neutral, or even {\em negative} influence on the likelihood of a rule, based upon the sign of $\theta_i$.

\section{System training and usage}
\label{sec:learning-parameters}

We are now ready to describe in full the operation of the training and inference phases.

\subsection{Training phase: learning $\theta$}
\label{sec:bootstrap}

At training time, we wish to learn the parameter $\theta$ that characterizes the conditional probability of a program given the input, $\Pr[f | z; \theta]$. We assume each training example $z^{(t)}$ is also annotated with the ``correct'' program $f^{(t)}$ that explains both the example and actual data pairs. We may attempt to discover these annotations automatically by bootstrapping: we start with a uniform parameter estimate $\theta^{(j)}=0$.  In iteration $j=1,2,3, \ldots$, we select $f^{(j,t)}$ to be the most likely program, based on $\theta^{(j - 1)}$, consistent with the system data.  (If no program is found within the timeout, the example is ignored.)  Then, parameters $\theta^{(j)}$ are learned, as described below.  This is run until convergence.

Fix a single iteration $j$. For notational convenience, we write target programs $f^{(t)}=f^{(j,t)}$ and parameters $\theta = \theta^{(j)}$.   We choose $\theta$ so as to minimize the negative log-likelihood of the data, plus a regularization term:
$$\theta = \underset{\theta' \in \mathbb{R}^n}{\operatorname{argmin}} -\log \Pr[f^{(t)}| z^{(t)}; \theta'] + \lambda \Omega(\theta'),$$
where $\Pr[f^{(t)}| z^{(t)}; \theta]$ is defined by equations (\ref{eqn:PCFG}) and (\ref{eqn:product}), the regularizer $\Omega(\theta)$ is the $\ell_2$ norm $\frac{1}{2} ||\theta||_2^2$, and $\lambda > 0$ is the regularization strength which may be chosen by cross-validation. If $f^{(t)}$ consists of rules $r^{(t)}_{1},r^{(t)}_{2},\ldots,r^{(t)}_{k^{(t)}}$ (possibly with repetition), then
$$\log \Pr[f^{(t)}| z^{(t)}; \theta] = \sum_{k=1}^{k=k^{(t)}}  \log \left(Z_{\mathrm{LHS(r^{(t)}_k)}}\right)
 - \!\!\! \sum_{i: r^{(t)}_k \in c_i(z^{(t)})} \!\!\! \theta_i$$
The convexity of the objective follows from the convexity of the regularizer and the log-sum-exp function. The parameters $\theta$ are optimized by gradient descent.

\subsection{Inference phase: evaluating on new input}

At inference time, we are given system input $z = (x, \xM, \yM)$, $n$ clues $c_1,c_2,\ldots,c_n$, and parameters $\theta\in \mathbb{R}^n$ learned from the training phase. We are also given a timeout $\tau$.  The goal is to infer the most likely program $\hat{f}$ that explains the data under a certain PCFG.  This is done as follows:
\renewcommand{\labelenumi}{(\roman{enumi})}
\vspace{-2mm}
\begin{enumerate}
	\item We evaluate each clue on the system input $z$. The underlying PCFG $\mathcal{G}_z$ consists of the union of all suggested rules, $\mathcal{R}_z=\bigcup_{i=1}^n c_i(z)$.
	\vspace{-1mm}
	\item Probabilities are assigned to these rules via Equation \ref{eqn:product}, using the learned parameters $\theta$.
	\vspace{-1mm}
	\item We enumerate over $\mathcal{G}_z$ in order of decreasing probability, and return the first discovered $\hat{f}$ that explains the $(\xM, \yM)$ string transformation, or $\perp$ if we exceed the timeout.
\end{enumerate}
\vspace{-2mm}
\renewcommand{\labelenumi}{(\arabic{enumi})}

To find the most likely consistent program, we enumerate all programs of probability at least $\eta>0$, for any given $\eta$.  We begin with a large $\eta$, gradually decreasing it and testing all programs until we find one which outputs $\yM$ on $\xM$ (or we exceed the timeout $\tau$). (If more than one consistent program is found, we just select the most likely one.)  Due to the exponentially increasing nature of the number of programs, this decreasing threshold approach imposes a negligible overhead due to redundancy -- the vast majority of programs are executed just once.

To compute all programs of probability at least $\eta$, a dynamic program first computes the maximal probability of a full trace from each nonterminal. Given these values, it is simple to compute the maximal probability completion of any partial trace. We then iterate over each nonterminal expansion, checking whether applying it can lead to any programs above the threshold; if so, we recurse.

\section{Results on prototype system}
\label{sec:expts}

To test the efficacy of our proposed system, we report results on a prototype web app implemented using client-side {JavaScript} and executed in a web browser on an Intel Core i7 920 processor. Our goal with the experiments is \emph{not} to claim that our prototype system is ``better'' than existing systems in terms of functionality or richness. ({Even if we wished to compare functionality, this would be difficult since all existing text processing systems that we are aware of are proprietary}.) Instead, our aim is to evaluate whether learning weights using textual features -- which has not been studied in any prior system, to our knowledge -- can speed up inference. Nonetheless, we do attempt to construct a reasonably functional system so that our results can be indicative of what we might expect to see in a real-world text processing system.

\subsection{Details of base functions and clues}

As discussed in Section \ref{sec:pcfg}, we associated the rules in our PCFG with a set of base functions. In total we created around $100$ functions, such as {\small\tt dedup}, {\small\tt concatLists}, and {\small\tt count}, as described in Section \ref{sec:example-operation}. For clues to connect these functions to features of the examples, we had one set of base clues that suggested functions we believed to be common, regardless of the system input $z$ (e.g.\ string concatenation). Other clues were designed to support common formats that we expected, such as dates, tabular and delimited data. Table \ref{tbl:clues} gives a sample of some of the clues in our system, in the form of grammar rules that certain textual features are connected to; in total we had approximately $100$ clues. The full list of functions and clues is available as part of our supplementary material.

\begin{table}[!h]
	\vspace{-4mm}
	\caption{Sample of clues used. {\tt LIST} denotes a list-, {\tt E} a string-nonterminal.}
	\centering
{\small
	\begin{tabular}{|ll|}
	\hline
	\textbf{Feature} & \textbf{Suggested rule(s)} \\
	\hline
	Substring $s$ appears in output but not input? & {$\mathtt{E \to ``s"}$, $\mathtt{LIST \to \{ E \}}$} \\ \hline
	Duplicates in input but not output? & {$\mathtt{LIST \to dedup(LIST)}$} \\ \hline	
	Numbers on each input line but not output line? & {$\mathtt{LIST \to count(LIST)}$} \\ 
	\hline
	\end{tabular}
}	
	\label{tbl:clues}
	\vspace{-2mm}
\end{table}

\subsection{Training set for learning}

To evaluate the system, we compiled a set of $280$ examples with both an example pair $(\xM,\yM)$ and evaluation pair $(x,y)$ specified. These examples were partly hand-crafted, based on various common usage scenarios the authors have encountered, and partly based on examples used in \citep{Gulwani}. All examples are expressible as (possibly deep) compositions of our base functions; the median depth of composition on most examples is around $4$. Like any classical learning model, we assume these are iid samples from the distribution of interest, namely over ``natural' text processing examples. It is hard to justify this independence assumption in our case, but we are not aware of a good solution to this problem in general; even examples collected from a user study, say, will tend to be biased in some way. Table \ref{tbl:test-cases} gives a sample of some of the scenarios we tested the system on. To encourage future research on this problem, our suite of training examples is ready for public release, and	is available as part of our supplementary material.

\begin{table}[!h]
\vspace{-3mm}
\caption{Sample of test-cases used to evaluate the system.}
	\centering
{\small
	\begin{tabular}{|p{2.5in}p{2.5in}|}
	\hline
	\textbf{Input} & \textbf{Output} \\
	\hline
	{\tt Adam Ant$\backslash$n1 Ray Rd.$\backslash$nMA$\backslash$n90113} & {\tt 90113} \\ \hline
	{\tt 28/6/2010} & {\tt June the 28th 2010} \\ \hline
	{\tt 612 Australia} & {\tt case 612: return Australia;} \\ \hline
	\end{tabular}
}
	
	\label{tbl:test-cases}
\end{table}

\subsection{Does learning help?}

The learning procedure aims to allow us to find the correct program in the shortest amount of time. We compare this method to a baseline, hoping to see quantifiable improvements in performance.

\textbf{Baseline}. Our baseline is to search through the grammar in order of increasing program size, attempting to find the shortest grammar derivation that explains the transformation. The grammar does use clues to winnow down the set of relevant rules, but does not use learned weights: we let $\theta_i = 0$ for all $i$, i.e.\ all rules that are suggested by a clue have the same constant probability. This method's performance lets us measure the impact of learning. 
{Note that pure brute force search would not even use clues to narrow down the set of feasible grammar rules, and so would perform strictly worse. Such a method is infeasible for the tasks we consider, because some of them involve e.g.\ constant strings, which cannot be enumerated.}

\textbf{Measuring performance}. To assess a method, we look at its \emph{accuracy}, as measured by the fraction of correctly discovered programs, and \emph{efficiency}, as measured by the {time} required for inference. As every target program in the training set is expressible as a composition of our base functions, there are two ways in which we might fail to infer the correct program: (a) the program is not discoverable within the timeout set for the search, or (b) another program (one which also explains the example transformation) is wrongly given a higher probability. We call errors of type (a) \emph{timeout errors}, and errors of type (b) \emph{ranking errors}. Larger timeouts lead to fewer timeout errors.

\textbf{Evaluation scheme}. One possible pitfall in an empirical evaluation is having an overly specific set of clues for the training set: an extreme case would be a single clue for each training example, which automatically suggested the correct rules to compose. To ensure that the system is capable of making useful predictions on new data, we report the test error after creating $10$ random $80$--$20$ splits of the training set. For each split, we compare the various methods as the inference timeout $\tau$ varies from $\{ 1/16, 1/8, \ldots, 16 \}$ seconds. For the learning method, we performed $3$ bootstrap iterations (see Section \ref{sec:bootstrap}) with a timeout of $8$ seconds to get annotations for each training example.

\begin{figure}[!t]
	\centering
	\subfigure[Timeout errors.]{\label{fig:test_error_timeout}\includegraphics[scale=0.2]{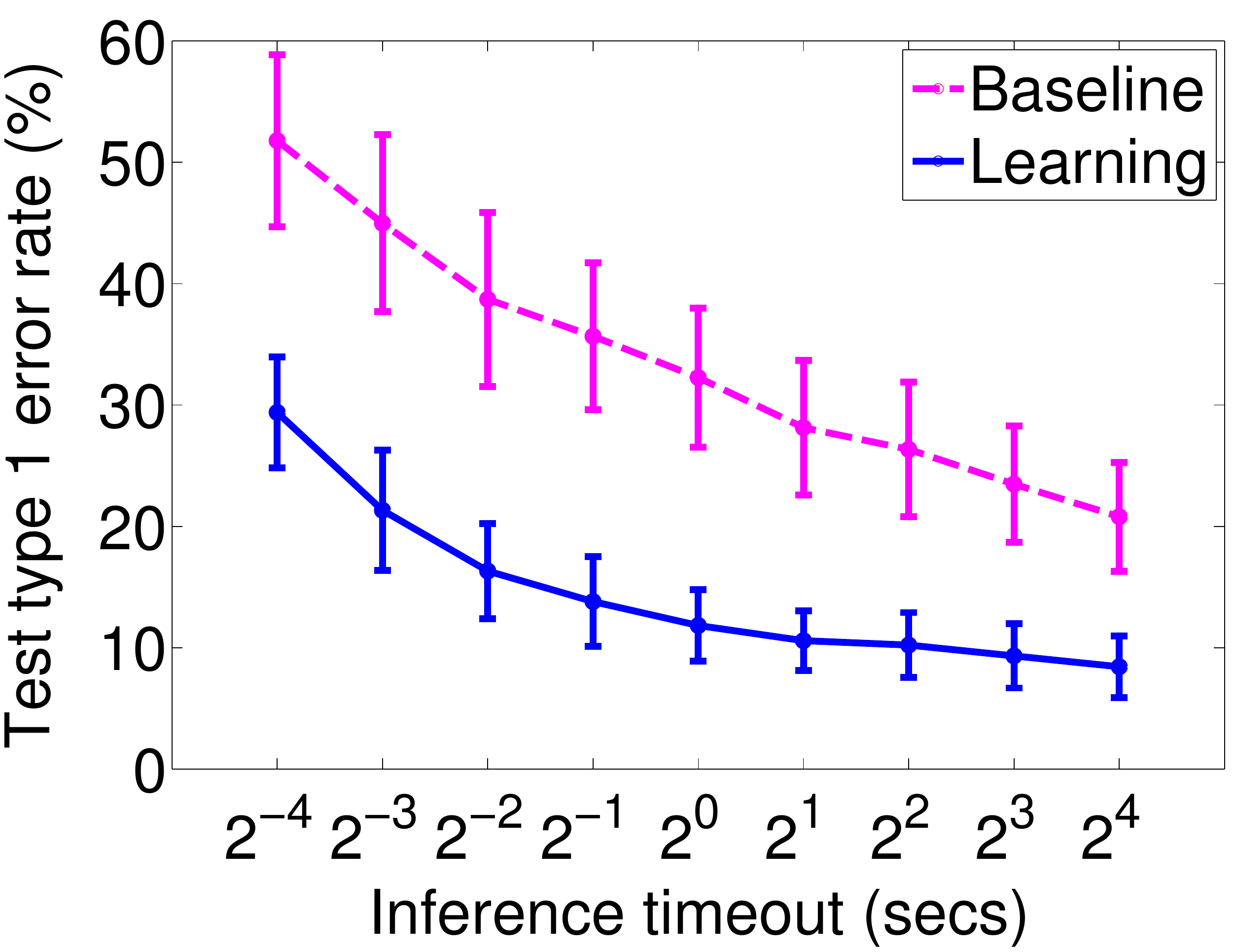}}\hspace{0.25in}
	\subfigure[Ranking errors.]{\label{fig:test_error_rank}\includegraphics[scale=0.2]{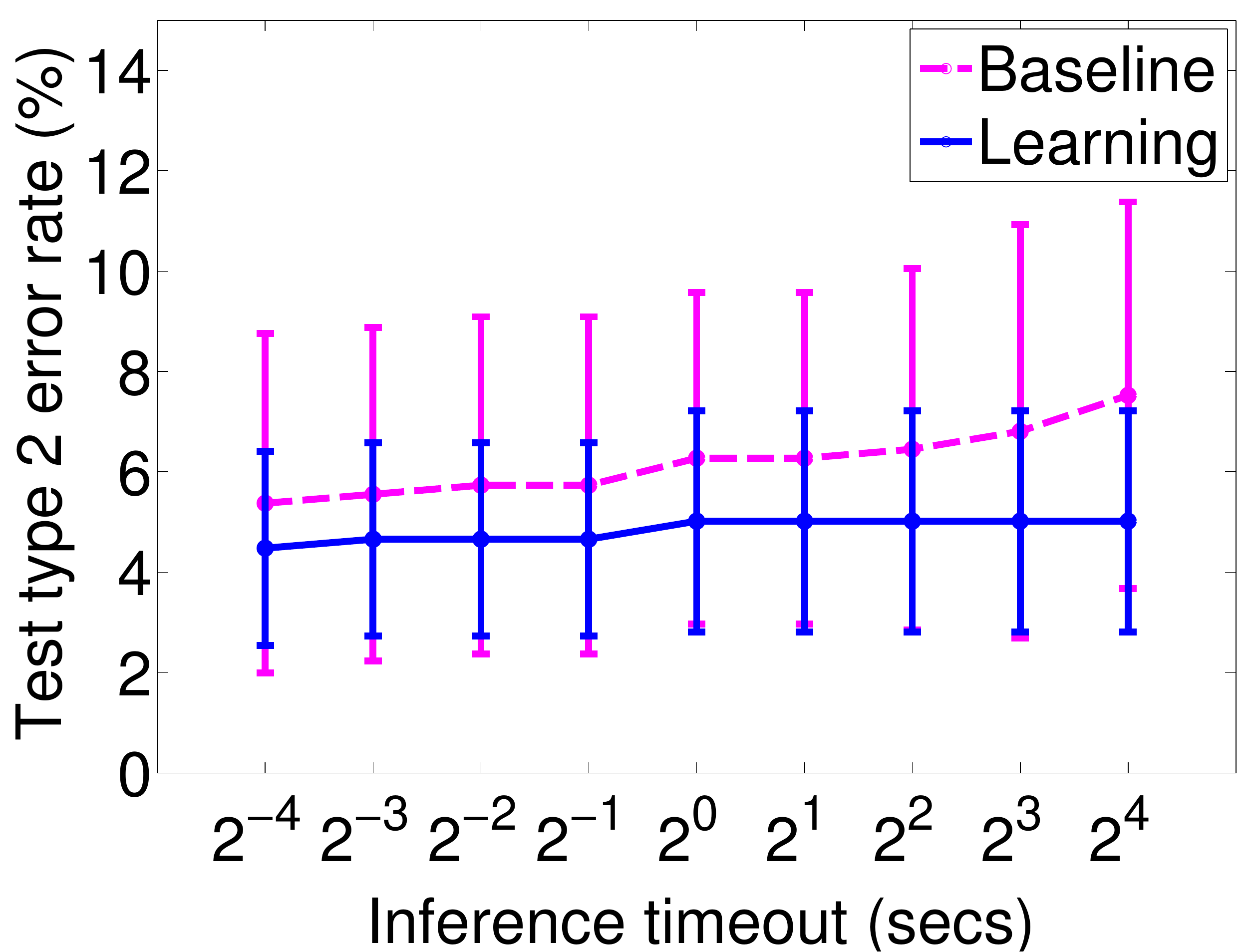}}
	\subfigure[Mean speedup due to learning.]{\label{fig:test_times_speed}\includegraphics[scale=0.2]{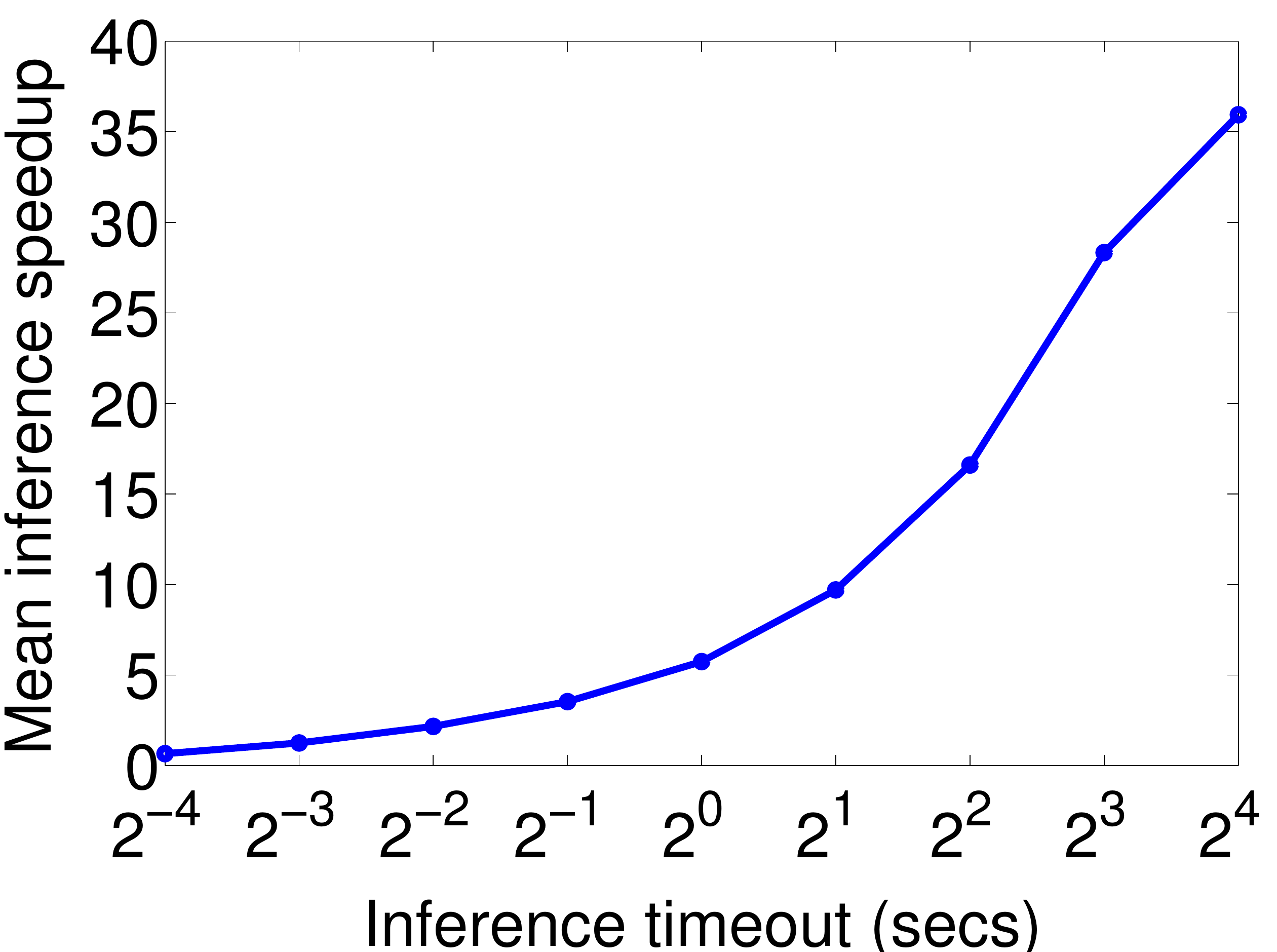}}\hspace{0.25in}
	\subfigure[Scatterplot of prediction times.]{\label{fig:test_times_scatter}\includegraphics[scale=0.2]{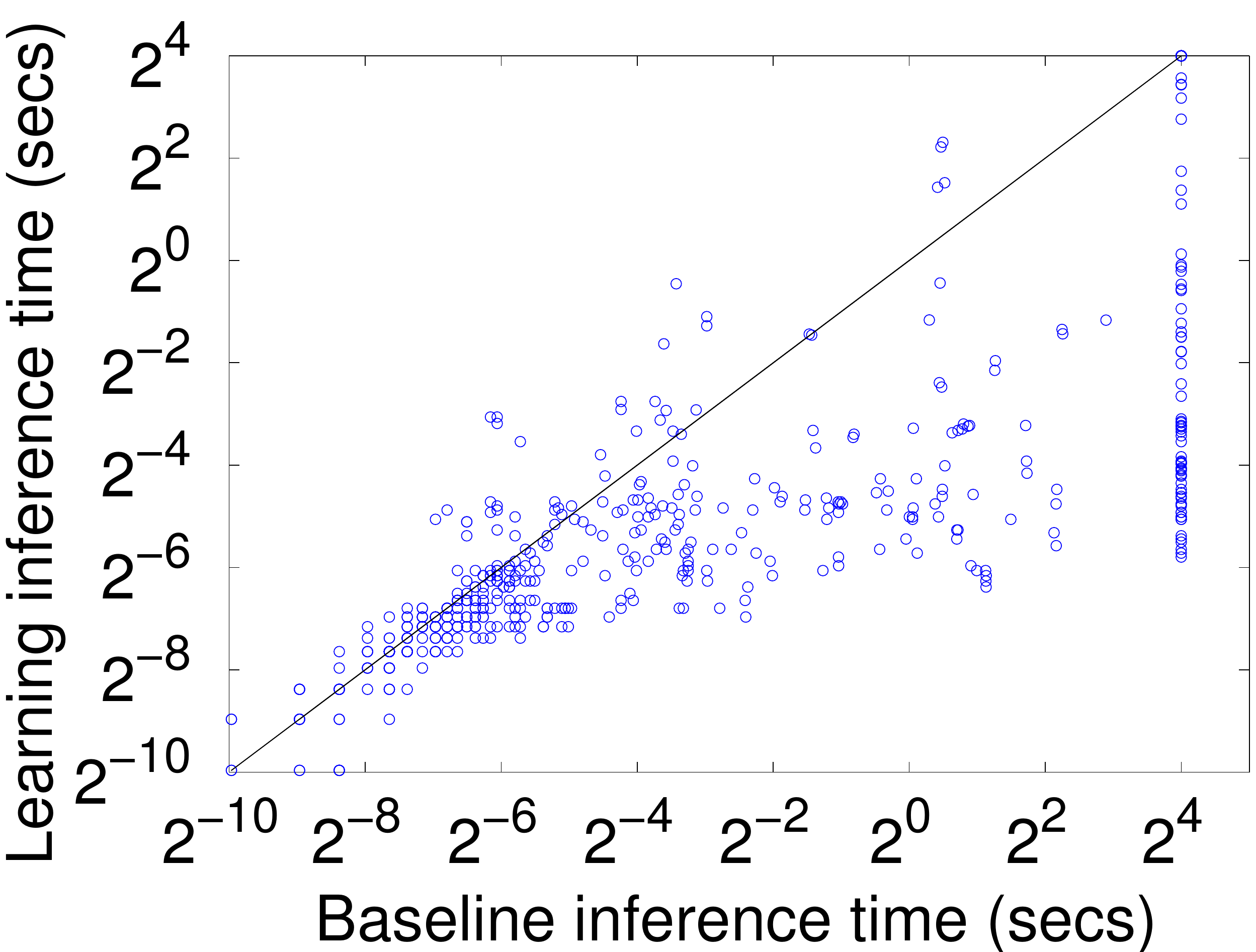}}
	\caption{Comparison of baseline versus learning approach.}	
	\vspace{-2mm}
\end{figure}

\textbf{Results}. Figures \ref{fig:test_error_timeout} and \ref{fig:test_error_rank} plot the timeout and ranking error rates respectively. As expected, for both methods, most errors arise due to timeout when the $\tau$ is small. To achieve the same timeout error rate, learning saves about two orders of magnitude in $\tau$ compared to the baseline. Learning also achieves lower mean ranking error, but this difference is not as pronounced as for timeout errors. This is not surprising, because the baseline generally finds few candidates in the first place (recall that the ranking error is only measured on examples that do not timeout); by contrast, the learning method opens the space of plausible candidates, but introduces a risk of some of them being incorrect.

Figure \ref{fig:test_times_speed} shows the relative speedup due to learning as $\tau$ varies. We see that learning manages to cut down the prediction time by a factor of almost $40$ over the baseline with $\tau = 16$ seconds. (The system would be even faster if implemented in a low-level programming language such as {C} instead of {Javascript}.) The trend of the curve suggests there are examples that the baseline is unable to discover with $16$ seconds, but learning discovers with far fewer. Figure \ref{fig:test_times_scatter} is a scatterplot of the times taken for both methods with $\tau = 16$ over all $10$ train-test splits, confirms this: in the majority of cases, learning finds a solution in much less time than the baseline, and solves many examples the baseline fails on within a fraction of a second. (In some cases, learning slightly increases inference time. Here, the test example involves functions insufficiently represented in the training set.)

\begin{figure}[!t]
	\centering
	{\includegraphics[scale=0.1825]{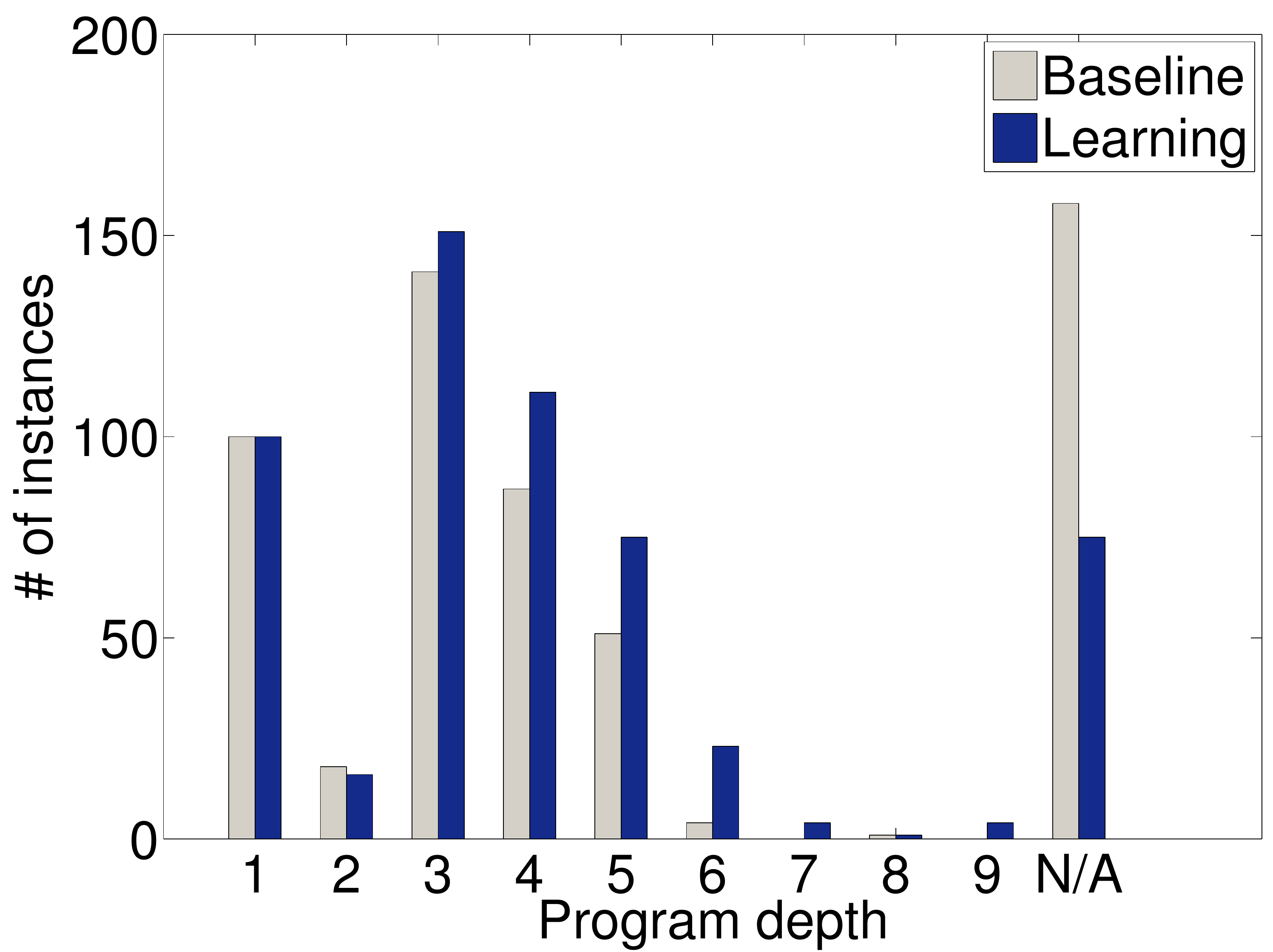}}
	\caption{Learnt program depths, $\tau = 16$s. ``N/A'' denotes that no successful program is found.}
	\label{fig:program_depths}
	\vspace{-2mm}
\end{figure}

Finally, Figure \ref{fig:program_depths} compares the depths of programs (i.e.\ number of constituent grammar rules) discovered by learning and the baseline over all $10$ train-test splits, with an inference timeout of $\tau = 16$ seconds. As expected, the learning procedure discovers many more programs that involve deep (depth $\geq 4$) compositions of rules, since the rules that are relevant are given higher probability.

\section{Conclusion and future work}

We propose a PBE system for repetitive text processing based on exploiting certain clues in the input data. We show how one can learn the utility of clues, which relate textual features to rules in a context free grammar. This allows us to speed up the search process, and obtain a meaningful ranking over programs. Experiments on a prototype system show that learning with clues brings significant savings over na\"{i}ve brute force search. As future work, it would be interesting to learn correlations between rules and clues that did \emph{not} suggest them, although this would necessitate enforcing some strong parameter sparsity. It would also be interesting to incorporate ideas like adaptor grammars \citep{Adaptor} and learning program structure as in \citep{Liang}.




\bibliography{bib/references}
\bibliographystyle{plainnat}

\end{document}